\documentclass[letterpaper]{article} %
\usepackage{aaai23}  %
\usepackage{times}  %
\usepackage{helvet}  %
\usepackage{courier}  %
\usepackage[hyphens]{url}  %
\usepackage{graphicx} %
\usepackage{natbib}  %
\usepackage{caption} %
\usepackage{algorithm}
\usepackage{algorithmic}

\usepackage{xspace}
\usepackage{newfloat}
\usepackage{listings}
\DeclareCaptionStyle{ruled}{labelfont=normalfont,labelsep=colon,strut=off} %
\floatstyle{ruled}
\newfloat{listing}{tb}{lst}{}
\floatname{listing}{Listing}
\usepackage{times}
\usepackage{latexsym}
\usepackage{xspace}
\usepackage[T1]{fontenc}
\usepackage{xcolor}
\usepackage[utf8]{inputenc}

\usepackage{microtype}
\usepackage{fancyhdr}
\usepackage{color}
\usepackage{url}
\usepackage{multirow}
\usepackage{graphicx} %
\graphicspath{{fig/}} %

\usepackage{placeins}

\usepackage{flushend}
\usepackage{balance}

\usepackage{colortbl}

\newcommand{\rise}[1]{\textcolor{gray}{\textsubscript{+#1}}}

\newcommand{\RiseB}[1]{\textcolor{blue}{\textsubscript{\bf +#1}}}

\newcommand{\bandc}{\cellcolor{gray!20}}
\newcommand\ExtraSep
{\dimexpr\cmidrulewidth+\aboverulesep+\belowrulesep\relax}

\usepackage{todonotes}
\newcommand{\anote}[1]{}

\newcommand{\fix}[1]{}

\newcommand{\TODO}[1]{}
\newcommand{\ignore}[1]{}

\newcommand{\here}[1]{}

\newcommand{\qnote}[1]{}

\newcommand{\hnote}[1]{}

\newcommand{\ynote}[1]{}

\newcommand{\methodName}{\textsc{WiGraph}\xspace}

\newcommand{\method}{\textsc{WiGraph}\xspace}

\newcommand{\mloss}{VIB-WI loss\xspace}

\newcommand{\vmask}{\textsc{VMASK}\xspace}
\newcommand{\base}{\textsc{BASE}\xspace}
\newcommand{\methodA}{\textsc{WIGRAPH-A}\xspace}
\newcommand{\methodAR}{\textsc{WIGRAPH-A-R}\xspace}

\newcommand{\methodnA}{\textsc{WIGRAPH-noA}\xspace}

\usepackage{booktabs}

\usepackage[nolist,nohyperlinks]{acronym}


\usepackage{lipsum}
\usepackage{titlesec}

\usepackage{tabularx} %
\usepackage{amssymb}%
\usepackage{amsmath}
\usepackage{mathtools}
\usepackage{pifont}%

\usepackage{ragged2e}

\usepackage{url}
\usepackage{dirtytalk}

\newcommand{\sref}[1]{Section~(\ref{#1})} 
\newcommand{\eref}[1]{Eq.~(\ref{#1})}

\newcommand{\cref}[1]{Condition~(\ref{#1})}

\include{macros}
\def\e{{\mathbf e}} 
\def\x{{\mathbf x}} 
\def\z{{\mathbf z}} 
\def\y{{\mathbf y}} 
\def\X{{\mathbf X}} 
\def\bZ{{\mathbf Z}} 
\def\Y{{\mathbf Y}}

\def\W{{\mathbf W}} 
 
\def\A{{\mathbf A}} 
\def\E{{\mathbf E}} 
\def\R{{\mathbf R}}

\usepackage{amsfonts,amssymb,bm}
\usepackage{pifont} %
\usepackage{graphicx,subfigure,epsfig,fancybox} %
\usepackage{float}
\usepackage{soul}
\usepackage{color} %
\usepackage{xcolor}
\usepackage{multirow}
\usepackage{natbib}
\newcommand{\hlc}[2][yellow]{{%
		\colorlet{foo}{#1}%
		\sethlcolor{foo}\hl{#2}}%
}

\newcommand{\revised}[1]{\textcolor{blue}{}}

\newcommand{\diag}[1]{\text{diag}\left(#1\right)}

\title{Improving Interpretability via Explicit \\ Word Interaction Graph Layer}

\author{Arshdeep Sekhon, Hanjie Chen, Aman Shrivastava, Zhe Wang, Yangfeng Ji, Yanjun Qi 
}

\affiliations {
    University of Virginia, Charlottesville, USA\\
    as5cu@virginia.edu, hc9mx@virginia.edu, as3ek@virginia.edu, zw6sg@virginia.edu, yj3fs@virginia.edu, yanjun@virginia.edu
}
\date{}

\begin{document}
\maketitle

\begin{abstract}
Recent NLP literature has seen growing interest in improving model interpretability. Along this direction, we propose a trainable neural network layer that learns a global interaction graph between words and then selects more informative words using the learned word interactions. Our layer, we call \method, can plug into any neural network-based NLP text classifiers right after its word embedding layer \footnote{code: \texttt{https://github.com/QData/WIGRAPH}}. Across multiple SOTA NLP models and various NLP datasets, we demonstrate that adding the \method layer substantially improves NLP models' interpretability and enhances models' prediction performance at the same time.
\end{abstract}

\section{Introduction}

Deep neural networks (DNNs) have achieved remarkable results in the field of natural language processing (NLP) \cite{zhang2015character,miwa2016end,wu2016google,wolf2020transformers}. Trustworthy real-world deployment of NLP models requires models to be not only accurate but also interpretable~\cite{xie2020explainable}. Literature has included a growing focus on providing posthoc explanations or rationales for NLP models' predictions \cite{ribeiro2016should,lundberg2017unified, murdoch2018beyond,  singh2018hierarchical,chen2020generating}. However, explaining DNNs using a posthoc manner cannot improve a model's intrinsic interpretability.

As shown in \cite{chen2020learning}, two NLP models may have the same prediction behavior but different interpretation ability. This concept of "intrinsic interpretability" motivates a compelling research direction to improve the interpretability of NLP models. A few recent studies used user-specified priors as domain knowledge to guide model training  \cite{camburu2018snli,du2019attribution,chen2019improving,erion2019learning,molnar2019quantifying}, hence improving model interpretability. Such information priors, however, may not be available in many tasks. Several other studies 
proposed to develop inherently interpretable models \cite{alvarez2018towards,rudin2019stop}, but these require intensive engineering efforts. More recently, \citet{chen2020learning} proposed to add a variational word mask, VMASK, to improve the interpretability of NLP neural classifiers.

The aforementioned literature on improving NLP models' intrinsic interpretability have mostly focused on highlighting important words. Strategies like VMASK  just select important words unilaterally, without accounting for how one word influences other words regarding interpretability. Studies have shown that word interactions are critical in explaining how NLP models make decisions~\cite{halford2010relational}.  For instance, for a sentiment classification task, when without a context, it is hard to conclude if the word {\it `different'} by itself is vital for sentiment? However, if we find `different' highly relates to  the word `refreshingly', it will likely contribute substantially to the model's sentiment prediction (see Table~\ref{tab:intro_examples} \footnote{These attribution interpretations were generated by LIME \citep{ribeiro2016should} and use BERT-base model on SST-2 dataset to explain two models' predictions.}).

\begin{table}
  \small
  \centering
  \begin{tabular}{p{0.1\textwidth}p{0.35\textwidth}}
    \toprule
    Model & ~~~~~~ Explanation \\
    \midrule
     \base &  still , this \hlc[pink!90]{thing} feels \hlc[pink!100]{flimsy} \hlc[pink!80]{and} ephemeral  \\
    \rule{0pt}{3ex} 
     \method &  still , this \hlc[cyan!90]{thing} feels \hlc[cyan!100]{flimsy} and \hlc[cyan!80]{ephemeral } \\
    \midrule
    \base & so \hlc[pink!100]{young} , so \hlc[pink!80]{smart} , \hlc[pink!70]{such} talent , such \hlc[pink!90]{a} wise \\\rule{0pt}{3ex} 
 \method &  so \hlc[cyan!90]{young} , so \hlc[cyan!100]{smart} , such talent , such \hlc[cyan!80]{a} \hlc[cyan!70]{wise}\\

   \midrule
    \base & it \hlc[pink!80]{is} risky , intelligent , \hlc[pink!100]{romantic} and \hlc[pink!70]{rapturous} \hlc[pink!90]{from} start to finish
    \\\rule{0pt}{3ex} 
  \method &  it is risky , \hlc[cyan!90]{intelligent} , \hlc[cyan!100]{romantic} and \hlc[cyan!70]{rapturous} \hlc[cyan!80]{from} start to finish \\
   
   \midrule
    \base & take \hlc[pink!80]{care} of my cat \hlc[pink!100]{offers} \hlc[pink!90]{a} refreshingly different slice of \hlc[pink!70]{asian} cinema
    \\\rule{0pt}{3ex} 
  \method &  take care of my cat \hlc[cyan!100]{offers} \hlc[cyan!90]{a} \hlc[cyan!70]{refreshingly} \hlc[cyan!80]{different} slice of asian cinema \\

    \bottomrule
  \end{tabular}
  \caption{Top ranked important words are shown in pink for \base and blue for \method augmented \base model. We can tell that word attributions from \method augmented model are easier to understand, and highlight more relevant sentiment words. This indicates \method augmented model has better intrinsic interpretibility than \base.}
    \vspace{2mm}
  \label{tab:intro_examples}
    \vspace{2mm}
\end{table}

Along this direction, we propose a novel neural network layer, we call \method, to improve NLP models' intrinsic interpretability.  \method is a plug-and-play layer and uses a graph-oriented neural network design: (1) It includes a stochastic edge discovery  module that can discover significant interaction relations between words for a target prediction task; (2) It then uses a neural message passing module to update word representations by using information from their interacting words; and (3) It designs a variational information bottleneck based loss objective to suppress irrelevant word interactions (regarding target predictions). We call such a loss: \mloss. To improve a target  text classifier's intrinsic interpretability, we propose to add the proposed \method layer right after the word embedding layer and fine-tune such an augmented model using a combination of the original objective and \mloss objective.

In summary, this paper makes the following contributions:
\begin{itemize}%
    \item We design \method  to augment neural text classifiers to improve these models'  intrinsic interpretability. \method does not require external user priors or domain knowledge. \method can plug-and-play into any  neural NLP models' architectures, right after the word embedding layer.  
    \item We provide extensive empirical results showing that adding \method layer into SOTA neural text classifiers  results in better explanations (locally, globally as well as regarding interactions)  and better model predictions  at the same time.   
\end{itemize}

\begin{figure*}
    \centering
    \includegraphics[width=\textwidth]{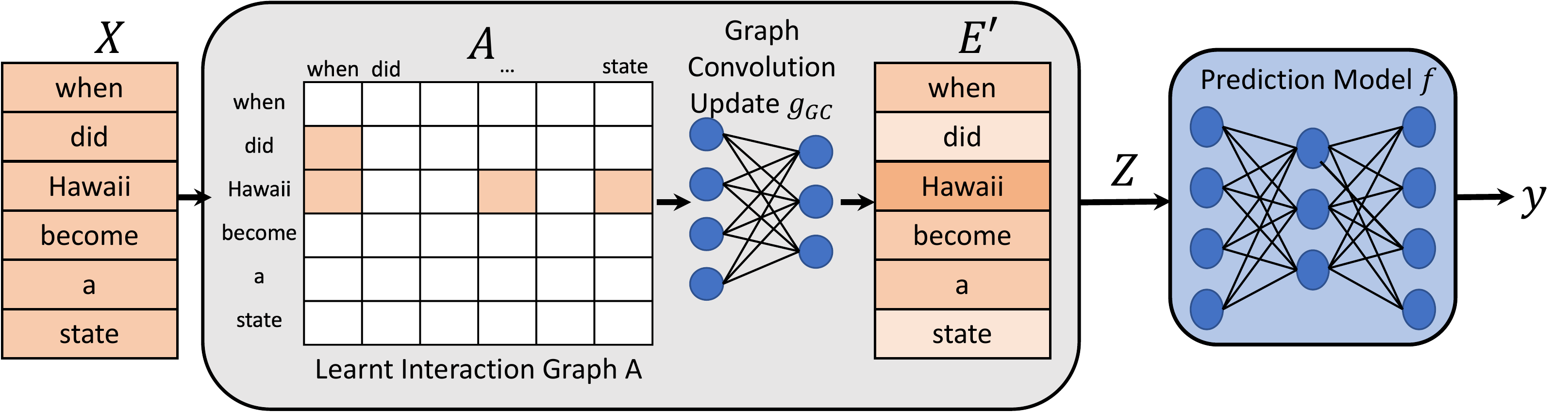}
    \caption{\methodName layer (components inside the gray box): during inference,  embeddings of  words (for example, {\it Hawaii} and {\it state}) are aggregated based on their interactions using a modified Graph Convolutional operation.  Here graph $\mathbf{A}$ was learnt from training along with the prediction task. A \method layer is inserted into a neural text classifier right after the word embedding input layer. }
    \label{fig:imask_model}
\end{figure*}

\section{Method: A Novel \method Layer}
\label{sec:method}

Our main hypothesis is: a novel layer that can extract crucial global word interactions will improve neural text classifiers' interpretability. This is because we envision plugging such a layer will enhance a target model's decision-making process by providing explicit guidance on what words are more important using the information on those words they interact with.  We aim for three properties in such a layer design: (1) plug-and-play; (2) model agnostic; and (3) no loss of prediction performance.

We denote vectors using lowercase bold symbols. %
We assume text inputs include a maximum length of $L$ tokens. We denote the whole word vocabulary as set $\mathbb{V}$. We use $V$ to denote its size (the total number of unique words in this vocabulary  $\mathbb{V}$). Besides, we use $\emph{f}$ to describe a neural text classification model. $\emph{f}$ classifies an input text into $\y \in \{1, \dots, C\}$, where $C$ is the number of classes. For an input sentence, we denote its $i$-th word as $w_i$ and its embedding representation as vector $\x_{i}$: $\forall i \in \{1, \dots, L\}$. Therefore, the embeddings of an input text make a matrix form  $\mathbf{X}=[\x_1, \dots, \x_L]^T$ (this means $\X \in \mathbb{R}^{L \times d}$ where $L$ is the length of input and $d$ is the dimension of each $\x_i$.).

\subsection{To Discover Word Interaction Graph: $A$}
\label{sec:A}

Now we explain the first component of the proposed \method layer. This module aims to discover how words globally interact for a predictive task. Our primary strategy to describe how words relate is to treat words as nodes and their interaction as edges in an interaction graph. We follow such an idea and choose to learn an undirected word interaction graph using a stochastic neural network module. We represent this unknown graph as $\mathbf{A} = \{ \mathbf{A}_{ij} \}_{V \times V}$. $\mathbf{A}$ includes the edges representing word interactions.  We assume each $\mathbf{A}_{ij} \in \{0,1\}$ is one binary random variable. $\mathbf{A}$ is stochastic whose $\mathbf{A}_{ij}$ specifies the presence or absence of an interaction between word $i$ and word $j$ in vocabulary $\mathbb{V}$.
 $\mathbf{A}_{ij} \in \{0,1\} $ is sampled from $ \mathit{Sigmoid}(\mathbf{\gamma}_{ij})$, following Bernoulli distribution with parameter $\mathit{Sigmoid}(\mathbf{\gamma}_{ij})$.   In Section~\ref{sec:gcn}, we show how $\mathbf{A}$ can help us understand how certain words are more important than others owing to the learned word interactions.

Learning the word interaction graph $\mathbf{A}$ means to learn the parameter matrix $\mathbf{\gamma}=\{ \mathbf{\gamma}_{ij} \}_{V \times V}$.  In Section~\ref{subsec:ibloss}, we  show how $\mathbf{\gamma}$ (and therefore $\mathbf{A}$) is learned  through the variational information bottleneck framework\cite{alemi2016deep}.

\subsection{Message Passing on Word Interaction Graph using Graph Convolution: $\mathbf{E}'$}
\label{sec:gcn}

 In our second module, we represent the $i$-th word $\x_i$ of an input text $\x$ as a node on the $\A$ graph. We  use a modified version of graph convolutional operation \cite{kipf2016semi} to update each $\mathbf{x}_i$ with its neighboring words $\mathbf{x}_j$. Here $j \in \mathcal{N}(i)$, and $\mathcal{N}(i)$ denotes those neighbor nodes of $\x_i$ on the graph $\mathbf{A}$ and in $\x$. Specifically, we denote the resulting word representation vector as $\e'_i$. Each $\mathbf{x}_i$ is revised using a graph based summation from its neighbors' embedding $\mathbf{x}_j, j \in \mathcal{N}(i)$:
\begin{equation}
    \e_i' = \x_i +  \sigma\Bigg( \tfrac{1}{|\mathcal{N}(i)|}\sum_{j \in \mathcal{N}(i)} \x_j \Bigg),
    \label{eq:GCN_vec}
\end{equation}
  Eq.~(\ref{eq:GCN_vec}) is motivated by the design of Graph convolutional networks (GCNs) that were recently introduced to learn useful node representations that encode both node-level features and relationships between connected nodes \cite{kipf2016semi}.  Different from the ReLU activation function used in vanilla GCNs, we use GeLU as the $\sigma(\cdot)$ , the non-linear activation function proposed in \cite{hendrycks2016gaussian}.

We want to point out that Eq.~(\ref{eq:GCN_vec}) is different from a typical GCN operation from \cite{kipf2016semi}. First, we only conduct one hop of neighbor aggregation in Eq.~(\ref{eq:GCN_vec}). A typical GCN module does  multi-hops. Second, we drop $\mathbf{W}^t \in \mathbb{R}^{d \times d}$ used for $t$-th hop of GCN update in \cite{kipf2016semi}. This is because we assume that the \base text classifier model $f$ has taken into account this prior and our \method layer will not bias to prefer short range interactions.  The third difference is the most important distinction that differentiates ours apart from \cite{kipf2016semi}. The graph has been given apriori to typical GCNs.  However, in our work, we need to learn the graph $\A$ (see Section~\ref{subsec:ibloss} on how to learn $\A$).

We can compute the simultaneous update of all words in input text $\x$ together by concatenating all $\e'_i$. This gives us one matrix $\E' \in \mathbb{R}^{L \times d}$, where $L$ is the length of input and $d$ is the embedding dimension of each $\x_i$. The simultaneous update can be written as:
\begin{equation}
\label{eq:roull_gcn}
\E' = \sigma(\mathbf{A}'\mathbf{X}).
\end{equation}
where $\mathbf{A'} = \mathbf{\hat{D}}^{-\frac{1}{2}}(\mathbf{A}_{\x} + \mathbf{I})\mathbf{\hat{D}}^{-\frac{1}{2}}$, that is the normalized adjacency matrix and $\mathbf{\hat{D}}$ is the diagonal degree matrix of $(\mathbf{A}_{\x}+ \mathbf{I})$. Note: $\mathbf{A}_{\x}$ denotes those edges from  $\mathbf{A}$ that are local for the current sample text $\x$. In summary, our second module computes: $$\mathbf{E'}=g_{GC}(\mathbf{X},\mathbf{A})$$

\fix{justify why harder to deal with multiple different A, compare with ElMo}

\subsection{To Build and Use \method Layer: $\mathbf{X}  \rightarrow \mathbf{Z}$ }

Our design of \method layer is that it can take the embedding matrix of a text example as input ($\X \in \mathbb{R}^{L \times d}$), and output a revised matrix representing each word with revised embedding ($\bZ \in \mathbb{R}^{L \times d}$).  \method layer aids the selection of more informative words based on their interactions for current predictive task. 
The proposed layer does not need significant efforts on engineering network architectures and does not require pre-collected importance attributions or explanations.

Our main goal is to improve the intrinsic interpretability of neural text classifiers with a simple model augmentation. Therefore, for a given neural text classifier, we propose to simply insert a \method layer right after the word embedding input layer and before the subsequent network layers of that target model.

There exist many possible ways to build \method layer from our first two modules (\sref{sec:A} and \sref{sec:gcn}). The simplest way is that we can just pass  $\E'$ as $\mathbf{Z}$. 

\begin{equation}
    \bZ = \E'
\end{equation}

Figure~\ref{fig:imask_model} visualizes how this vanilla version of \method layer updates word representations with the  $\mathbf{X}  \rightarrow  \E' \rightarrow \mathbf{Z}$ data flow during inference. During training, it needs to learn the graph $\A$.

\subsection{Model Training with \mloss}
   \label{subsec:ibloss}

Now we propose to train \method jointly with other layers using a new objective that we name as variational information bottleneck loss for word interaction (\mloss). \mloss aims to restrict the information of globally irrelevant word interactions flowing to subsequent network layers, hence forcing the model to focus on important interactions to make predictions. Following the Information Bottleneck framework \cite{alemi2016deep}, we aim to learn $\A$ and all subsequent layers' weights $\{\W\}$, to make $\mathbf{Z}$ maximally informative of the prediction label ${\Y}$, while being maximally compressive of $\mathbf{X}$ (see Figure~\ref{fig:imask_model}). That is 

    \begin{equation}
      max_{\A, \{\W\}} \{ I(\mathbf{Z};\mathbf{Y}) - \beta I(\mathbf{Z}; \mathbf{X}) \} 
        \label{eq:ib_main_objective}
    \end{equation}  
Here $I(\cdot;\cdot)$ denotes the mutual information, and $\beta\in\mathbb{R}_{+}$ is a coefficient balancing the two mutual information terms.

Given a specific example $(\x^{m},\y^{m})$, we can further simplify the lower bound of first term $I(\mathbf{Z};\mathbf{Y})$ in \eref{eq:ib_main_objective} as: 
\begin{equation}
 \label{eq:mvib-1}
I(\z;\y^{m}) \geq \mathbb{E}_{q(\z|\x^{m})} log (p(\y^m| \x^m; \A, \{\W\}))
\end{equation}

Similarly for the second term $I(\mathbf{Z}; \mathbf{X})$ in \eref{eq:ib_main_objective} and for a given example $(\x^{m},\y^{m})$, we can simplify its upper bound as: 
\begin{eqnarray}
\label{eq:mvib-2}
I(\z;\x^m)\leq  KL(q(\mathbf{A} | \x^m) || p_{a0}(\mathbf{A}))
\end{eqnarray}
Due to the difficulty in calculating two mutual information terms in \eref{eq:ib_main_objective}, we follow~\cite{schulz2020restricting,alemi2016deep} to use a variational approximation $q(\X,\Y,\bZ)$ to approximate the true distribution $p(\X,\Y,\bZ)$. Details on how to derive \eref{eq:mvib-1} and \eref{eq:mvib-2} are in \sref{sec:moreIBloss}. Now combining \eref{eq:mvib-1} and \eref{eq:mvib-2} into \eref{eq:ib_main_objective}, we get the revised objective as:   
\begin{equation}
\label{eq:mlossa}
\begin{aligned}
 max_{\A, \R, \{\W\}} \{ \mathbb{E}_{q(\mathbf{Z}|\x^m)} log (p(\mathbf{y}^m| \x^m; \A, \R, \{\W\})) \\
   - \beta_g KL(q(\mathbf{A} | \x^m) || p_{a0}(\mathbf{A})) \}
   \end{aligned}
\end{equation}
\eref{eq:mlossa} is the proposed VIB objective for a given observation $(\x^{m},\y^{m})$.

\paragraph{Detailed Model Specification: }
During training, for the stochastic interaction graph $\A$, we learn its trainable parameter matrix $\gamma \in \mathbb{R}^{|V| \times |V|}$, that is also optimized along with the model parameters
during training. Further, we use the mean field approximation \cite{blei2017variational}, that is, $q(\mathbf{A}_{\x}|\x) = \prod_{i=1}^L \prod_{j=1}^L q(A_{x_i, x_j}|\x_i,\x_j)$.

 Equation~\ref{eq:mlossa} requires prespecified prior distributions $p_{a0}$. We use a Bernoulli distribution prior (a non-informative prior) for each word-pair interaction $q_{\phi}[\mathbf{A}_{x_i, x_j}|\x_i,\x_j ]$. $p_{a0}(\mathbf{A}_{x})=\prod_{i=1}^L\prod_{j=1}^L p_{a0}(\mathbf{A}_{\x_i, \x_j})$ and $p_{a0}(\mathbf{A}_{x_i, x_j})= Bernoulli(0.5)$.  This leads to:
\begin{equation}
   KL(q(\mathbf{A}_{\x} | \x^m) || p_{a0}(\mathbf{A})) = -H_{\mathbf{q}} (\mathbf{A}_{\x}| \x^m) 
\end{equation}
Here, $H_q$ denotes the entropy of the term $\mathbf{A}_{\x}| \x^m$ under the $q$ distribution. 
Besides, we add a sparsity regularization on $\mathbf{A}_{\x}$ to encourage learning of sparse interactions. Now, we have the following loss function for $(\x^{m},\y^{m})$:
\begin{equation}
\begin{split} 
 - (\mathbb{E}_{\x} p(\y|\x^m;  \mathbf{A}, \{\W\}) + 
    \beta_{g} H_{\mathbf{q}}(\mathbf{A}_{\x}| \x^m)) \\ + \beta_{sparse} ||\mathbf{A}_{\x}||_1 
    \end{split}
    \vspace{-5mm}
\end{equation}
 During training,  $\A$ is discrete and is drawn from Bernoulli distributions that are parametrized with  matrix $\gamma \in \mathbb{R}^{|V| \times |V|}$. We, therefore, use the Gumbel-Softmax\cite{jang2016categorical} trick to differentiate through the sampling step and propagate the gradients to the respective parameters $\gamma$.

\subsection{Variation: \method-A-R} 

We also try another possible design of \method that includes one more separate module that learns an attribution word mask $\R$ on top of $\E'$.  Aiming for better word selection, $\R$ is designed as a stochastic layer we need to learn and $\R \in \{0,1\}^{V}$. Each entry in $\R$ (e.g., $\R_j \in\{0,1\}$) follows a Bernoulli distribution with parameter $\phi$ (to be learned).

During inference, for an input text $\x$, we get a binary vector $\R_{\x}$ from $\R$ that is of size $L$. Its $i$-th entry $\R_{\x_i}\in\{0,1\}$ is a binary random variable associated with the word token at the $i$-th position. %
We use the following operation (a masking operation!) to generate the final representation of the $i$-th word from a \method layer:   
\begin{equation}
\z_i = \R_{\x_i} \e'_i 
\end{equation}

We can compute the simultaneous update of all words in input text $\x$ together by concatenating all $\z_i$ denoted as matrix $\bZ \in \mathbb{R}^{L \times d}$. The simultaneous update can then be written as: 
\begin{equation}
\label{eq:wmask}
\bZ = \diag {\R_{\x}}_{L \times L} \E'_{L \times d}
\end{equation}
During training, we need to learn both $\A$ and $\R$. Now the loss function \mloss turns to:
\begin{equation*}
\begin{split} 
 - (\mathbb{E}_{\x} p(\y|\x^m;  \mathbf{A}, \mathbf{R}, \{\W\}) + \beta_i H_q(\mathbf{R}_{\x}|\x^m) + \\
    \beta_{g} H_{\mathbf{q}}(\mathbf{A}_{\x}| \x^m)) + \beta_{sparse} ||\mathbf{A}_{\x}||_1 
    \end{split}
\end{equation*}
Due to page limit, we put detailed derivations of above and specification of $\R$ in \sref{subsec:moreibloss}. %
We call the vanilla version of \method as \textbf{\method-A} and the version with the word mask R as \method-A-R. 
\section{Connecting to Related Work}

Our design orients from one basic notion that we treat interpretability as an intrinsic property of neural network models. We expect a neural text classifier will be more interpretable, when focusing on important word interactions to make predictions. Our work connects to multiple related topics: 

\paragraph{Self-Explaining Models} Recent literature has seen growing interests in treating interpretability as an {\it inherent property} of NLP deep models. \cite{alvarezmelis2018robust,rudin2019stop} proposed to design self-interpretable models by requiring human annotations in model engineering. \cite{chen2020learning} proposed VMASK layer for improving NLP models' interpretability. This layer automatically learns task-specific word importance and guides a  model to make predictions based on important words. However, this method does not consider interactions between words. 

\paragraph{Explanations as Feedback} 
Anoter category of work uses explanations as feedback for improving model prediction performance as well as to encourage explanation faithfulness. 
\cite{camburu2018snli,chen2019improving,erion2019learning,molnar2019quantifying} focuses on aligning human judgments with generated explanations and further incorporating it into the training of the model. These methods require human annotations that are expensive to obtain and also have the risk of not aligning well with the separately trained model's decision making process.  \cite{ross2017right,ross2017improving,rieger2020interpretations} use explanations as feedback into the model to improve prediction performance. However, these heavily rely on  ground-truth explanations and domain knowledge. Differently, our proposed method augments a model with a special layer that improves both prediction performance and interpretability (see \sref{sec:exp}).

\paragraph{Graph Neural Networks}  Graph Neural Networks (GNNs) generalize neural networks from regular grids, like images  to irregular structures like graphs.  There exists a wide variety of GNN architectures like  \cite{kipf2016semi,scarselli2008graph,velivckovic2017graph,santoro2017simple}. They share the same underlying concept of message passing between connected nodes in the graph.  However, little attention has been paid to address cases when the underlying  graph is unknown.   In contrast, in \methodName, we do not know the global interaction graph {\it apriori}. It is learnt along with the prediction model as part of the training.

\paragraph{Post-Hoc Explanation} NLP literature includes a number of methods that focus on disentangling the rationales of a trained NLP model's decision by finding which words contributed most to a prediction, including the popularly used LIME\cite{ribeiro2016should} and SampleShapley \cite{kononenko2010efficient} methods. 
Recent studies have proposed to generate post-hoc explanations beyond word-level features by detecting feature interactions, including for instance, contextual decomposition by \cite{murdoch2018beyond}. Other work adopted Shapley interaction index to compute feature interactions \cite{lundberg2018consistent}. In contrast to these post-hoc interpretation systems, our method focuses on  designing a strategy to improve the inherent interpretability of NLP models.

\begin{table*}[!h]
\begin{minipage}[]{\hsize}
	\centering
\scalebox{0.89}{
    \begin{tabular}{llllllll}
		\toprule
		\base & Models & IMDB & SST-1 & SST-2 &  AG News & TREC & Subj \\
		\midrule
		 \multirow{3}{*}{LSTM}	&  \base  & 88.39 &  43.84 &  83.74 &  91.03 &  90.40 &  90.20 \\
		\rule{0pt}{2ex}  
			& \vmask   & 90.07  & 44.12  & 84.35  &  92.19  & 90.80  & 91.20 \\
		\rule{0pt}{2ex}
		& \bandc \method  & \bandc {\bf 90.12} \RiseB{1.73} & \bandc {\bf 46.47} \RiseB{2.63} & \bandc {\bf 86.21} \RiseB{2.63}  &\bandc {\bf 91.16} \rise{0.13} &\bandc {\bf 92.20} \RiseB{1.80}  & \bandc {\bf 91.40} \RiseB{1.20}     \\

	\midrule
	
	\multirow{3}{*}{BERT}	&  \base   &   91.88 &  51.63 &  92.15  & 92.05  &  97.40 & 96.40 \\	
		& \vmask   & 93.04  & 51.36  & 92.26  &  94.24  & 97.00 &  96.40 \\\rule{0pt}{2ex}
		
			& \bandc\method  & \bandc {\bf 92.48} \RiseB{0.60}  & \bandc {\bf 52.49} \RiseB{0.86} &  \bandc {\bf 92.59} \rise{0.44} & \bandc {\bf 92.72} \RiseB{0.67} & \bandc {\bf 97.40} \rise{0.00} & \bandc {\bf 96.60} \rise{0.20} \\
		
			\midrule
	
	\multirow{3}{*}{RoBERTa}	&  \base   & 89.87 & 55.20 & 94.73 & 93.46 & 96.2 & 96.00 \\	
		    &  \vmask    & 90.02 &  54.21 & 93.47  & 93.47 & 96.0 & 96.50 \\
			& \bandc\method    & \bandc {\bf 90.10} \rise{0.23} & \bandc {\bf 55.52} \rise{0.32}  &\bandc {\bf 94.75} \rise{0.02} & \bandc {\bf 93.52} \rise{0.06} & \bandc {\bf 96.60}\rise{0.20} & \bandc {\bf 96.40} \rise{0.40}  \\
	
		\midrule
	\multirow{3}{*}{distilBERT}	&  \base   & 86.96 & 51.31 & 90.50 & 93.34 & 97.20 & 96.20 \\	
		    &  \vmask    &  87.00 & 48.01 & 89.02 & 93.81 & 95.20 & 95.00 \\
			& \bandc \method    & \bandc {\bf 88.32} \RiseB{1.36}  &\bandc 50.81  &\bandc {\bf 90.77} \rise{0.22} &\bandc {\bf 93.85} \RiseB{0.51} &\bandc {\bf 97.40} \rise{0.20} & \bandc  {\bf 96.30}  \rise{0.10} \\
			\bottomrule 
	\end{tabular}}
	\caption{Prediction Accuracy (\%). Models augmented with \method layer predict better than \base. }
	\label{tab:prediction_acc}
	\end{minipage} 

\end{table*}

\begin{table}[h]
\begin{minipage}[]{.91\hsize}
\centering
\scalebox{0.9}{
    \begin{tabular}{ccccc}\toprule
       Dataset  & Train/Dev/Test & C & V  & L \\\hline
       sst1  &  8544/1101/2210 & 5 &  17838  & 50 \\\hline
       sst2  & 6920/872/1821 & 2 &  16190  & 50 \\\hline
       imdb  & 20K/5K/25K & 2  &  29571  & 250 \\\hline
       AG News & 114K/6K/7.6K & 4 &  21838  & 50 \\\hline
        TREC & 5000/452/500 & 6 & 8026  & 15 \\\hline
        Subj & 8000/1000/1000 & 2 & 9965  & 25 \\\bottomrule
    \end{tabular}}
    \caption{Summary of datasets we use in experiments:  number of classes ($C$),  vocabulary size ($V$) and sentence length ($L$). }
    \label{tab:dataset_stats}
\end{minipage} 
\end{table}

\paragraph{Information Bottleneck Based Methods}
The information bottleneck method was first proposed by \cite{tishby2000information,tishby2015deep}. \cite{alemi2016deep} introduced a variational approximation to the information bottleneck that enables usage for deep neural networks.  \cite{schulz2020restricting,bang2019explaining} utilized the information bottleneck principle to generate post-hoc explanations by highlighting important features while suppressing unimportant ones. Differently, we incorporate the information bottleneck in model training to make model prediction behavior more interpretable.

\begin{table*}[!ht]
	\centering
\scalebox{0.8}{\begin{tabular}{p{30mm}|cccccccc}
		\toprule
		Metrics & \base &  Models & IMDB & SST-1 & SST-2 &  AG News & TREC & Subj \\
		\midrule
		
				\multirow{6}{*}{\begin{tabular}{l}AOPCs \\[\ExtraSep] of LIME \\[\ExtraSep] Generated \\[\ExtraSep] Explanations \end{tabular}}  &
	
		\multirow{3}{*}{LSTM}	& \base & 14.34 &	8.76 &	17.03	& 7.00	& 11.95	& 9.67  \\
		\rule{0pt}{2ex}  
	& 	& \vmask  & 15.1 &	9.52 &	22.14 &	7.39 &	11.97 &	11.68 \\
		\rule{0pt}{2ex}  
		&	& \bandc \methodA & \bandc {\bf  17.8} &	\bandc {\bf 10.33} &\bandc {\bf 	22.34} &	\bandc {\bf 14.94} &	\bandc {\bf 20.13}	&\bandc  {\bf 16.27}  \\
		 \cline{2-9}
	&\multirow{3}{*}{BERT}	&   \base   & 10.63	& 26.08	& 43.96	& 7.12	&  {\bf 68.82}	& 44.13\\	\rule{0pt}{2ex}
		& & \vmask  &  {\bf 12.64} & 	27.5 & 	41.6 & 	8.47 & 	65.14 & 	44.41  \\	\rule{0pt}{2ex}
	&	& \bandc \methodA  & \bandc  10.96  &  \bandc {\bf	34.81}  & \bandc   {\bf	44.59}  &  \bandc  {\bf	11.13}  & \bandc 	68.51  & \bandc 	 {\bf 44.90} \\
	
	\cline{2-9}
	&\multirow{3}{*}{RoBERTa}	&   \base   &  {\bf 10.30} & 	32.00 & 	41.51 & 	15.7 & 	66.07 & 	43.00  \\	\rule{0pt}{2ex}
		& & \vmask  & 9.88 & 26.02 & 40.69 & 8.47 & 64.58 &  43.66  \\\rule{0pt}{2ex}
	&	& \bandc \methodA   &\bandc  10.23  & \bandc  {\bf	32.70}  & \bandc  {\bf	42.64}  & 	\bandc  {\bf 15.72}  & 	\bandc  {\bf 66.27}  & 	 \bandc {\bf 44.22}   \\
	\cline{2-9}
	&\multirow{3}{*}{distilBERT}	&   \base   & 11.00	 & 27.92	 & 42.25 & 	8.83 & 	67.63 &   {\bf 44.93} \\	\rule{0pt}{2ex}
		& & \vmask & 9.20		& 22.11		& 39.77		& 8.00	& 	63.14	& 	40.65\\	\rule{0pt}{2ex}
	&	& \bandc \methodA 	&  \bandc  {\bf 11.32}	&  \bandc  {\bf	36.21}	&  \bandc {\bf	44.33 }		&  \bandc {\bf 9.02}	& 	\bandc  {\bf 68.74}	& \bandc 	43.97 \\
	
		\midrule

		\multirow{6}{*}{\begin{tabular}{l}AOPCs of \\[\ExtraSep] SampleShapley \\[\ExtraSep] Generated \\[\ExtraSep] Explanations\end{tabular} } & 
		
	\multirow{3}{*}{LSTM}		&  \base &	15.80	&	7.91	&	22.38	&	6.62	&	11.90	&	11.66	\\\rule{0pt}{2ex} 
	&	& \vmask & 16.48	&	9.73	&	22.52	&	7.65	&	11.86	&	12.74 \\
		\rule{0pt}{2ex}  
	&	& \bandc	\methodA  & \bandc 	{\bf 21.73}	& \bandc  {\bf	9.78}	&\bandc  {\bf	49.40}	&\bandc 	{\bf 27.60}	& \bandc  {\bf	68.27}	&	\bandc {\bf 29.29}  \\
	\cline{2-9}
	& \multirow{3}{*}{BERT}	&\base  &	13.00	&	28.65	&	41.65	&	7.21	&	65.37	&	33.22\\
			\rule{0pt}{2ex}  
&		& \vmask &	12.18	&	29.92	&	41.53	&	10.02	&	65.14	&	44.41 \\
			\rule{0pt}{2ex}  
&		& \bandc 	\methodA  &\bandc  {\bf	14.16}	&	\bandc {\bf 36.02}	&	\bandc {\bf 44.75}	&\bandc 	{\bf 10.43}	&	\bandc {\bf 66.30}	&	\bandc {\bf 44.91} \\
\cline{2-9}
	 &\multirow{3}{*}{RoBERTa}	&   \base  & {\bf	9.13}	&	36.01	&	35.89	&	6.01	&	66.07	&	13.08 \\	\rule{0pt}{2ex}
		& & \vmask  &	8.00	&	29.89	&	42.64	&	5.68	&	54.85	&	{\bf 43.75}  \\	\rule{0pt}{2ex}
	&	& \bandc \methodA  & \bandc 	8.50	&	\bandc {\bf 38.01}	&\bandc 	{\bf 42.78}	&\bandc 	{\bf 7.12}	&\bandc 	{\bf 67.56}	&\bandc 	43.61 \\
	
	\cline{2-9}
	&\multirow{3}{*}{distilBERT}	&   \base   &	12.03	&	28.07	&	42.24	&	15.09	&	47.78	&	13.08\\	\rule{0pt}{2ex}
		& & \vmask &	9.26	&	18.67	&	35.03	&	13.95	&	59.53	&	40.93  \\	\rule{0pt}{2ex}
	&	& \bandc \methodA &\bandc {\bf	12.18}	&	\bandc {\bf 35.29}	&\bandc	{\bf 43.14}	& \bandc	{\bf 14.21}	& \bandc {\bf	66.08}	&	\bandc {\bf 43.97} \\
	\bottomrule
	\end{tabular}}
	\caption{AOPCs (\%) obtained from results using LIME and SampleShapley to interpret the base,  \methodName-based models and the baseline \vmask across four SOTA models, and over six different datasets. }
	\label{tab:aopc}
\end{table*}

\section{Experiments}
\label{sec:exp}
\vspace{-1mm}
We design experiments to answer the following: 
  \begin{enumerate}
\item Are NLP models augmented with \method layer more interpretable models? 
\item Do NLP models augmented with \method layer predict well?  
\end{enumerate}
Besides, we extend \method to one concept based vision task in Section~\ref{sec:concept}.

\subsection{Setup: Datasets, Models and Metrics}

\paragraph{Datasets} Our empirical analysis covers six popular text classification datasets as detailed by Table~\ref{tab:dataset_stats}. These six datasets are "sst1", "sst2"\cite{socher2013recursive}, "imdb"\cite{maas2011learning}, "AG news"\cite{Zhang2015CharacterlevelCN}, "TREC"\cite{li2002learning} and "Subj"\cite{pang2005seeing}. Three of the datasets are for binary classification, and the rest are for multi-class text classification tasks. 

\paragraph{\base Models} We use four commonly used neural text classifiers to evaluate \method: LSTM, and transformer based SOTA models including BERT, RoBERTa and distilBERT. As Section~\ref{subsec:ibloss}  described, we plug our \method layer right after the word embedding input layer. For the LSTM models\cite{hochreiter_long_1997}, we initialize word embeddings from \cite{mikolov2013distributed} with dimension $d=300$.  For BERT, RoBERTa and distilBERT models, we use fine-tuned base models from \cite{wolf2020transformers}  on each dataset (Table~\ref{tab:dataset_stats}).

\paragraph{Hyperparameter Tuning}  We perform fine-tuning on each model (batch size=64). We fix the word embedding layer and train \method layer along with the rest of a \base model.  For the LSTM models, we vary the hidden size $\in \{100,300,500\}$, and  dropout in $\{0.0,0.2, 0.3\}$. We set $\beta_{sparse} \in \{1e-02,1e-03, 1e-04\}$, $ \beta_g \in \{1.0, 1e-02,1e-03, 1e-04\}$ and $ \beta_i \in \{1.0, 1e-02,1e-03, 1e-04\}$. 
The learning rate is tuned from the set $\{0.0001, 0.0005, 0.005, 0.001\}$. For transformer based models, we vary dropout in range $\{0.2, 0.3, 0.5\}$, hidden dimension  to compute $\mathbf{R} \in \{128, 256, 512\}$. We set $\beta_{sparse}, \beta_g, \beta_i=1.0$ and anneal it by a factor of 0.1 every epoch.  For the larger vocabulary cases (IMDB, AG-News datasets and transformer-base models), we filter words for learning our interaction matrix $\mathbf{A}$, i.e., we learn interactions for the top frequent $10,000$ words. %

\paragraph{Baselines:} To our best knowledge, \method is the only plug-and-play layer to improve a target neural text classifier's interpretability using explicit pairwise word interactions. In our experiments, we compare \method to a \base model without \method layer and to a \base model augmented by the \vmask layer. \cite{chen2020learning} proposed  \vmask layer  for improving NLP models' intrinsic interpretability, though this layer does not consider word interactions.

\paragraph{Evaluation Metrics:} We use three types of evaluations to compare \method with baselines. (a) Prediction accuracy: this is to measure if NLP models augmented with \method layer predict well. (b) To compare different models' interpretability, we will apply two post-hoc attribution techniques:  LIME\cite{ribeiro2016should} and SampleShapley \cite{kononenko2010efficient} on model predictions. The resulting  feature attribution outputs will be evaluated using explanation faithfulness scores like AOPCs (details in \sref{sec:aopc}). (3) We further design  interaction interpretability measures to compare different models in \sref{sec:ios}.

\subsection{Prediction Performance Comparison}

In Table~\ref{tab:prediction_acc}, we compare prediction performance using four different SOTA models across six different datasets. This makes 24 different (\base, data) combinations , and on each case, we compare \base model, \vmask augmented base model versus our \method augmented model regarding the prediction accuracy. Here we refer to \method as the best performing model between \methodA and \methodAR.  Table~\ref{tab:prediction_acc} shows that adding \method layer into SOTA neural text classifier models makes the models predict better! Empirically, the performance gains on LSTM models appear more than on Transformer models.

\begin{table*}[h]
	\centering
\begin{minipage}[]{\hsize}
\centering
\scalebox{0.89}{
    \begin{tabular}{cccccccc}
		\toprule
		\base & Models & IMDB & SST-1 & SST-2 &  AG News & TREC & Subj \\
		\midrule
		 \multirow{2}{*}{LSTM}	&  \methodAR  & 89.12 & 44.32 & 84.08  &  {\bf 91.16}  & 91.65  & 90.60  \\
		\rule{0pt}{2ex}  
			& \methodA   & {\bf 90.12} &  {\bf 46.47}  &  {\bf 86.21 } & 90.87  &  {\bf 92.20}  &  {\bf 91.40}
		\\
\midrule

	\multirow{3}{*}{BERT}	&  \methodAR  & 90.81 &  {\bf 52.49}  &   {\bf 92.59} & 90.13  & 96.60  & 96.40   \\
		\rule{0pt}{2ex}  
			& \methodA   &  {\bf 92.48} &  52.04 &   91.54  &   {\bf 92.72} &  {\bf 97.40}  &  {\bf 96.60}
		\\
	\midrule
	\multirow{3}{*}{RoBERTa}	&  \methodAR  &  {\bf 90.10 } &  52.90 &  92.97  & 91.54  & 95.20  &  95.50 \\
		\rule{0pt}{2ex}  
			& \methodA   & 88.21 &  {\bf 54.52 } &  {\bf 94.45}  &   {\bf 93.52} &  {\bf 96.60}  &  {\bf 96.40}
		\\
	\midrule
	\multirow{3}{*}{distilBERT}	&  \methodAR  &  {\bf 88.32} &  {\bf 50.81 } & 88.19  &  {\bf 93.85}  & 96.40  &  96.20 \\
		\rule{0pt}{2ex}  
			& \methodA   & 88.07 &  49.95 &  {\bf 90.77 } & 91.08  &  {\bf 97.40} &  {\bf 96.30} \\
		\bottomrule
	\end{tabular}}
	\caption{Ablations analysis regarding prediction accuracy from: \methodA and \methodAR. }
	\label{tab:ablations}
	\end{minipage} 

\end{table*}

\subsection{Attribution Interpretability Comparison: Area Under Perturbation Curve (AOPC)}
\label{sec:aopc}
Here we empirically check our hypothesis that training a  model augmented with \method layer leads to improvements of model explanation faithfulness during  downstream post-hoc interpretation analyses.  We use Area Over Perturbation Curve (AOPC)~\cite{nguyen2018comparing,samek2016evaluating} as the evaluation metric. AOPC is defined as the {\it average change of prediction probability on the predicted class over a test dataset by deleting top $K$ words in explanations.} Higher AOPC scores reflect better interpretation faithfulness. \begin{equation}
    AOPC = \dfrac{1}{K+1} \sum_{k=1}^K <f(\x) - f(\x_{\backslash 1, \dots, k})>_{p(\x)} 
\end{equation}

We generate word-level attribution explanations using two popular post-hoc explanation methods: LIME\cite{ribeiro2016should} and SampleShapley \cite{kononenko2010efficient}. Across all datasets, we use $500$ test samples and  $k \in \{1, \dots, 10\}$. %
Table~\ref{tab:aopc} shows that \methodName-A outperforms the original \base model and the \base with \vmask. 

When LIME is used to post-hoc explain models, across all 24 cases of (model, dataset) combinations,  \methodName outperforms the original \base model and the \base with \vmask layer regarding AOPC score in 21 cases. The only three exception include the IMDB/BERT, TREC/BERT and IMDB/RoBERTa setups. When SampleShapley is used, \methodName outperforms the original \base model and the \base with \vmask layer in 22 cases out of 24  (model, dataset) combinations.

\subsection{Interaction  Analysis and Ablation}
\label{sec:ios}

In this experiment, %
we introduce a new metric: {\it Interaction Occlusion Score} (IoS). The IoS score  measures the interaction interpretability faithfulness of a target model on its learnt interactions.

\methodName discovers globally informative interactions with the importance score $\mathbb{E}_q[\mathbf{A}_{x_{i,j}}| \x_{i,j}]$ (see $q$ in \sref{subsec:ibloss}).  We sort entries of $\mathbf{A}$ and filter out the top $K$ global interaction scores, denoted by $\mathbf{A}_{ij}^k$.   We then calculate the accuracy of the model after only using these top $k$ interactions via:

\begin{equation}
\text{IOS}(k) = \dfrac{1}{M} \sum_{m=1}^M {1}_{{y}_m = y_m^k}
\end{equation}
Here, we represent the label of the model on the $m^{th}$ test sample as ${y}_m$. 
Table~\ref{tab:ios} in  ~\sref{sec:moreab}  shows using the top interactions outperforms the setting when no pairwise interactions are used during inference.

\paragraph{Ablation:} We perform extensive ablation analysis 
to  compare \methodA and \methodAR. Table~\ref{tab:ablations} provides comparison analysis regarding prediction accuracy and we can tell no clear winner between the two variations across all 24 cases of (\base, data) combinations. See \sref{sec:moreab}  and Table~\ref{tab:ablations_aopc} for more ablation results via other metrics. We recommend to use \methodA in most real-world applications, due to less parameters.   
\paragraph{Qualitative Visualization} \sref{sec:moreab} includes an extensive list of qualitative analyses we make to understand \method, including correlation analysis of interaction against co-occurrence, word clouds visualization, and  interacting word-pair examples. We show in Figure~\ref{fig:inter_freq_corr} that our learned interaction matrix does not merely mirror the co-occurrence statistics, instead learns more general informative global interactions.   Figure~\ref{fig:inter_cloud} uses word clouds to visualize top interacting word pairs obtained from TREC, SST1, and SST2 datasets.   The word pairs are consistent with  corresponding tasks. 

\subsection{Modeling Concept Interaction in Vision }
\label{sec:concept}

\citet{koh2020concept} introduced the notion of high-level concepts as an intermediate interpretable interface in the input-model-predictions pipeline. These concepts describe high level attributes of an image. This enables users to directly interact with a model by intervening on human-interpretable concepts. The interactions between these concepts can affect prediction, however these interactions are unknown.   In this section, we investigate the utility of learning these interactions for aiding prediction using our  \method layer. We train a concept bottleneck~\cite{koh2020concept} model on the CUB dataset~\cite{wah2011caltech} which is jointly trained with a \method layer. In Section~\ref{sec:conceptMore}, we show that the \method layer is able to learn concept embeddings, model interactions between concepts, and can also be used with test time concept intervention to improve prediction accuracy on the final task. In order to extend \method for this setup, we introduce a dynamic interaction graph with concepts and the image as nodes. Intuitively, the image is considered to be a composition of concept embeddings. The interaction between the concepts are learned variables whereas the interactions between an image and its concepts are computed using the cosine similarity. The details of our model architecture are in Section~\ref{sec:conceptMethod}.
\paragraph{Results:} As a baseline, we fine-tune an Inception V3-based joint concept bottleneck model~\cite{koh2020concept} that achieve a prediction accuracy of $80.04\%$ on the CUB dataset~\cite{wah2011caltech}. Together with this model, we jointly train our \method and the concept embedding layer to learn the interactions between the concepts. As described in~\cite{koh2020concept}, test time intervention (TTI) helps improve prediction accuracy to $89.41\%(+9.37\%)$. Interestingly, we observe that TTI achieves more improvements of prediction accuracy when using \method  augmented  concept vision model to $95.74\%(+15.70\%)$.

\section{Conclusions}
In this paper, we try to  answer the question:  \emph{Does adding a special layer in the form of discovering word-word interactions lead to improvements in a neural text classifier's interpretability?} Our paper gives a firm "Yes" to the question and provides a neural-network based design for making such a layer. The second component of \method layer uses the message passing framework and it can be expanded to  allow for learning and accounting for  higher-order interactions (not only pairwise) in a scalable way. We will explore this in our future works. Furthermore, \methodName can easily extend to cross sentence tasks like Natural Language Inference, and we will leave it to future.   %

\FloatBarrier
\clearpage

\bibliography{refArsh}

\FloatBarrier
\clearpage
\appendix
\section{Appendix}

\subsection{Model Training for \method-A-R}
   \label{subsec:moreibloss}

We propose to train \method jointly with other layers using a new objective that we name as variational information bottleneck loss for word interaction (\mloss). \mloss aims to restrict the information of globally irrelevant word interactions flowing to subsequent network layers, hence forcing the model to focus on important interactions to make predictions. Following the Information Bottleneck framework \cite{alemi2016deep}, we aim to learn $\A$, $\R$ and all subsequent layers' weights $\{\W\}$, to make $\mathbf{Z}$ maximally informative of ${\Y}$, while being maximally compressive of $\mathbf{X}$ (see Figure~\ref{fig:imask_model}). That is 
    \begin{equation}
      max_{\A, \R, \{\W\}} \{ I(\mathbf{Z};\mathbf{Y}) - \beta I(\mathbf{Z}; \mathbf{X}) \} 
        \label{eq:ib_main_objective_a}
    \end{equation}  
Here $I(\cdot;\cdot)$ denotes the mutual information, and $\beta\in\mathbb{R}_{+}$ is a coefficient balancing the two mutual information terms.

Given a specific example $(\x^{m},\y^{m})$, we can further simplify the lower bound of first term $I(\mathbf{Z};\mathbf{Y})$ in \eref{eq:ib_main_objective_a} as: 
\begin{equation}
 \label{eq:mvib-1-a}
I(\z;\y^{m}) \geq \mathbb{E}_{q(\z|\x^{m})} log (p(\y^m| \x^m; \A, \R, \{\W\}))
\end{equation}

Similarly for the second term $I(\mathbf{Z}; \mathbf{X})$ in \eref{eq:ib_main_objective_a} and for a given example $(\x^{m},\y^{m})$, we can simplify its upper bound as: 
\begin{eqnarray}
\label{eq:mvib-2-a}
I(\z;\x^m)\leq  KL(q(\mathbf{R} | \x^m) || p_{r0}(\mathbf{R}))\nonumber \\
   + KL(q(\mathbf{A} | \x^m) || p_{a0}(\mathbf{A}))
\end{eqnarray}
Due to the difficulty in calculating two mutual information terms in \eref{eq:ib_main_objective_a}, we follow~\cite{schulz2020restricting,alemi2016deep} to use a variational approximation $q(\X,\Y,\bZ)$ to approximate the true distribution $p(\X,\Y,\bZ)$. Details on how to derive \eref{eq:mvib-1-a} and \eref{eq:mvib-2-a} are in \sref{sec:moreIBloss}. Now combining \eref{eq:mvib-1-a} and \eref{eq:mvib-2-a} in \eref{eq:ib_main_objective_a}, we get the revised objective as:   
\begin{equation}
\label{eq:mlossa-a}
\begin{aligned}
 max_{\A, \R, \{\W\}} \{ \mathbb{E}_{q(\mathbf{Z}|\x^m)} log (p(\mathbf{y}^m| \x^m; \A, \R, \{\W\})) \\
   - \beta_i KL(q(\mathbf{R} | \x^m) || p_{r0}(\mathbf{R})) \\
   - \beta_g KL(q(\mathbf{A} | \x^m) || p_{a0}(\mathbf{A})) \}
   \end{aligned}
\end{equation}
\eref{eq:mlossa-a} is the proposed VIB objective for a given observation $(\x^{m},\y^{m})$. To increase the flexibility, we associate two different coefficients from $\mathbb{R}_{+}$ with the two KL-terms. In practice we treat them as hyper-parameters. %

\paragraph{Detailed Model Specification: }
In this section, we describe in detail how to learn discrete $\A$ and $\R$ along with the model parameters during training. To learn the word mask $\R$, we use amortized variational inference\cite{rezende2015variational}. That is, we use a single-layer feedforward neural network as the inference network $q_{\phi}(R_{x_t}| x_t)$, associated parameters $\phi$ are optimized with the model parameters
during training.  For the interaction mask (graph) $\A$, we use a trainable parameter matrix $\gamma \in \mathbb{R}^{|V| \times |V|}$, that is also optimized along with the model parameters
during training. Further, we use the mean field approximation \cite{blei2017variational} for both the word mask and the variational interaction mask, that is, $q(\mathbf{R}|\x) = \prod_{i=1}^L q(R_{x_t}|\x_t)$ and  $q(\mathbf{A}_{\x}|\x) = \prod_{i=1}^L \prod_{j=1}^L q(A_{x_i, x_j}|\x_i,\x_j)$.

 Equation~\ref{eq:mlossa-a} requires prespecified prior distributions $p_{r0}$ and $p_{a0}$. We use the Bernoulli distribution prior (a non-informative prior) for each word-pair interaction $q_{\phi}[\mathbf{A}_{x_i, x_j}|\x_i,\x_j ]$. $p_{a0}(\mathbf{A}_{x})=\prod_{i=1}^L\prod_{j=1}^L p_{a0}(\mathbf{A}_{\x_i, \x_j})$ and $p_{a0}(\mathbf{A}_{x_i, x_j})= Bernoulli(0.5)$.  This leads to:
\begin{equation}
   KL(q(\mathbf{A}_{\x} | \x^m) || p_{a0}(\mathbf{A})) = -H_{\mathbf{q}} (\mathbf{A}_{\x}| \x^m) 
\end{equation}
Here, $H_q$ denotes the entropy of the term $\mathbf{A}_{\x}| \x^m$ under the $q$ distribution. Similarly, for the word mask, $p_{r0}(\mathbf{R}) = \prod_{i=1}^L p_{r0}(\mathbf{R}_{\x_i})$, and $p_{r0}(\mathbf{R}_{\x_i})=Bernoulli(0.5)$. Therefore, 
\begin{equation}
    KL(q(\mathbf{R}_{\x} | \x^m) || p_{a0}(\mathbf{R})) = -  H_{\mathbf{q}} (\mathbf{R}_{\x}| \x^m)
\end{equation}

 Finally, we have the following loss function for $(\x^{m},\y^{m})$:
\begin{equation}
\begin{split} 
 - (\mathbb{E}_{\x} p(\y|\x^m;  \mathbf{A}, \mathbf{R}, \{\W\}) + \beta_i H_q(\mathbf{R}_{\x}|\x^m) + \\
    \beta_{g} H_{\mathbf{q}}(\mathbf{A}_{\x}| \x^m)) + \beta_{sparse} ||\mathbf{A}_{\x}||_1 
    \end{split}
    \vspace{-5mm}
\end{equation}
We use stochastic gradient descent to optimize the above loss using all training samples. During training, both $\A$ and $\R$ are discrete samples drawn from Bernoulli distributions in  ~\eref{eq:roull_gcn} and ~\eref{eq:wmask}. We, therefore, use the Gumbel-Softmax\cite{jang2016categorical} trick to differentiate through the sampling step and propagate the gradients to their respective parameters $\gamma$ and $\phi$.

\subsection{More results on ablations and Global Interaction Analysis results}
\label{sec:moreab}

We present our results on the IoS score proposed in Section~\ref{sec:ios} in Table~\ref{tab:ios}. We show using the
top interactions outperforms the setting when no pairwise interactions are used during inference. 

We perform further ablations to calculate the different proposed interpretability scores in Table ~\ref{tab:ablations_aopc}. We use LIME as the post-hoc explanation method for AOPC. \methodA achieves higher interpretability scores, both global and local, than its variations.

\begin{table*}[!h]
\begin{minipage}[]{\hsize}
	\centering
\scalebox{0.89}{	
    \begin{tabular}{ccllllll}
		\toprule
		Methods & Models & IMDB & SST-1 & SST-2 &  AG News & TREC & Subj \\
		\midrule

	\multirow{2}{*}{LSTM}	&  \methodnA   & 88.53 &  45.70 & 83.96  & {\bf 91.07 } & 91.00  & 90.30 \\
		\rule{0pt}{2ex}  
		
		& \bandc \method-topA  & \bandc {\bf 88.84 } & \bandc {\bf 45.82} & \bandc {\bf 84.48}  & \bandc 90.91 & \bandc  {\bf 91.27 } & \bandc {\bf 90.58}    \\
	\midrule
	\multirow{2}{*}{BERT}	&  
	\methodnA & 85.62  & 51.31 & {\bf 89.18}   & {\bf 90.79  } & 96.40  &  95.90 \\
	
			& \bandc  \method-topA & \bandc  {\bf 85.67} & \bandc {\bf 51.52} & \bandc 89.02  & \bandc 90.47 & \bandc {\bf 97.04 } &  \bandc {\bf 95.97}  \\
	
	\midrule
	\multirow{2}{*}{RoBERTa}	&  
	\methodnA & 89.02 & 52.10 & 91.52 & 90.13 & 95.2 & 95.50 \\
	
			& \bandc  \method-topA & \bandc {\bf 90.02 } &  \bandc {\bf 53.84 } &\bandc {\bf  92.51 } & \bandc {\bf 91.50 } & \bandc {\bf 96.00 } & \bandc {\bf 96.20 }\\
	
		\midrule
	\multirow{2}{*}{distilBERT}	&  
	\methodnA & 85.08 & 47.10 & 86.82 & 90.08 & 95.00 & 95.20 \\
	
			& \bandc  \method-topA & \bandc {\bf 86.32} & \bandc {\bf 47.25} & \bandc {\bf 87.00} & \bandc {\bf 90.10}  & \bandc {\bf 96.40} & \bandc {\bf 96.00 } \\
		
		\bottomrule
	\end{tabular}}
	\caption{Global Interaction Analysis: Average {\it Post Hoc Interaction Occlusion Score} when using top $K$ scoring interactions for prediction. We compare to \methodnA that sets $\A$ to identity matrix.}
	\label{tab:ios}
	\end{minipage} 
	\vspace{-2mm}
\end{table*}

\begin{table*}[th]
\begin{minipage}[]{.97\hsize}
	\centering
\scalebox{0.97}{	
    \begin{tabular}{cccccccc}
		\toprule
		Models & Method & IMDB & SST-1 & SST-2 &  AG News & TREC & Subj \\
		\midrule
		 \multirow{2}{*}{LSTM}	&  \vmask 	& 89.42		& 45.96		& 85.11	& 	92.00		& 93.48		& 92.50\\
		\rule{0pt}{2ex}  
			&  \methodAR 	&  88.2	& 	43.00		& 83.21	& 	91.78	& 	91.15	& 	91.60\\
		\rule{0pt}{2ex}  
			& \methodA  	&  88.12	& 	44.26	& 	84.78	& 	92.89	& 	93.03	& 	91.50
		\\
\midrule

	\multirow{3}{*}{BERT}	&  \vmask  & 92.90 & 51.95 & 92.32  & 93.72   & 96.68  & 98.00  \\
		\rule{0pt}{2ex}  
			&  \methodAR  & 90.50 & 52.77 &  92.43 &    90.10 &  96.68 &  97.90 \\
		\rule{0pt}{2ex}  
			& \methodA   & 91.88 & 50.50 & 92.89 & 92.00 &   96.02    & 97.60
		\\
	\midrule
	\multirow{3}{*}{RoBERTa}	&  \vmask  & 87.00 & 53.68  &  93.35 &  94.05  & 96.68  & 97.50  \\
		\rule{0pt}{2ex}  
			&  \methodAR  & 83.12 & 52.01 & 93.46   &  93.91 & 95.38  &   97.30 \\
		\rule{0pt}{2ex}  
			& \methodA   & 86.20 & 53.41 & 93.81  &  94.55  & 95.50  & 97.20
		\\
	\midrule
	\multirow{3}{*}{distilBERT}	&  \vmask  & 88.24 & 48.86 & 90.14  &  94.20  &  93.58 & 95.80  \\
		\rule{0pt}{2ex}  
			&  \methodAR  & 88.14 & 49.5 & 90.14  &  93.01  &  93.14 &   95.50 \\
		\rule{0pt}{2ex}  
			& \methodA   & 89.12 & 49.05 & 91.17  &   94.00 &  95.58 & 97.10 \\
		\bottomrule
	\end{tabular}}
	\caption{Validation Prediction Accuracy }
	\label{tab:ablations}
	\end{minipage} 

\end{table*}

\begin{table*}[h]
	\centering
\begin{minipage}[]{\hsize}
\centering
\scalebox{0.89}{
    \begin{tabular}{cccc}
		\toprule
		 Metric & Models & TREC & SST-2   \\
		\midrule 
	
	\multirow{2}{*}{IoS} & \methodAR-topA &  96.20 & { 86.90 } \\
		& \bandc \methodA-topA  & \bandc {\bf 96.40 } & \bandc {\bf 87.00 } \\	\midrule 
		\multirow{4}{*}{AOPC} & \base  & 67.63 & 42.25 \\	
		 & \vmask & 63.14 & 39.77 \\	
		 & \methodAR & 61.52 & 36.22 \\
		& \bandc \methodA & \bandc {\bf 68.74} & \bandc {\bf 44.33}\\
		\bottomrule
	\end{tabular}}
	\caption{Ablations analysis for TREC and SST-2 datasets on distilbert model regarding : IoS scores and AOPC from LIME post-hoc explanation model for ablations \methodA and \methodAR. }
	\label{tab:ablations_aopc}
	\end{minipage} 

\end{table*}
\subsection{More Qualitative Results}
\label{sec:qual}

\begin{figure}[!ht] 
  \centering 
    \begin{minipage}[]{0.47\textwidth}
      \centering 
      \includegraphics[width=1\columnwidth,height=40mm]{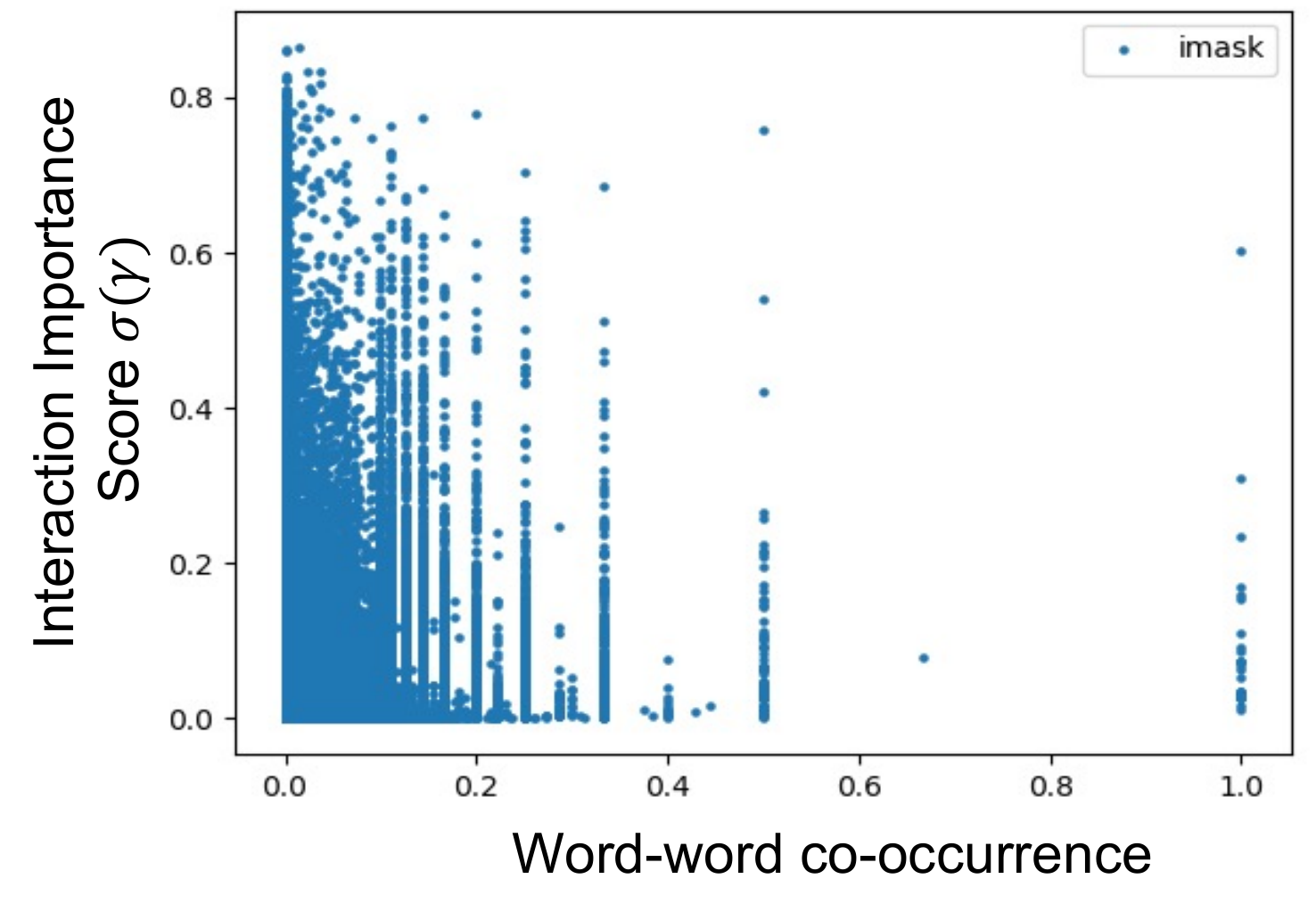}
    \end{minipage} 
    \caption{SST2 dataset with LSTM model. Interaction Importance Score vs word co-occurrence. \label{fig:inter_freq_corr}}
    \vspace{2mm}
\end{figure}

\begin{figure*}[htbp]
  \centering 
    \begin{minipage}[t]{.31\textwidth} 
      \centering 
        \includegraphics[width=\linewidth]{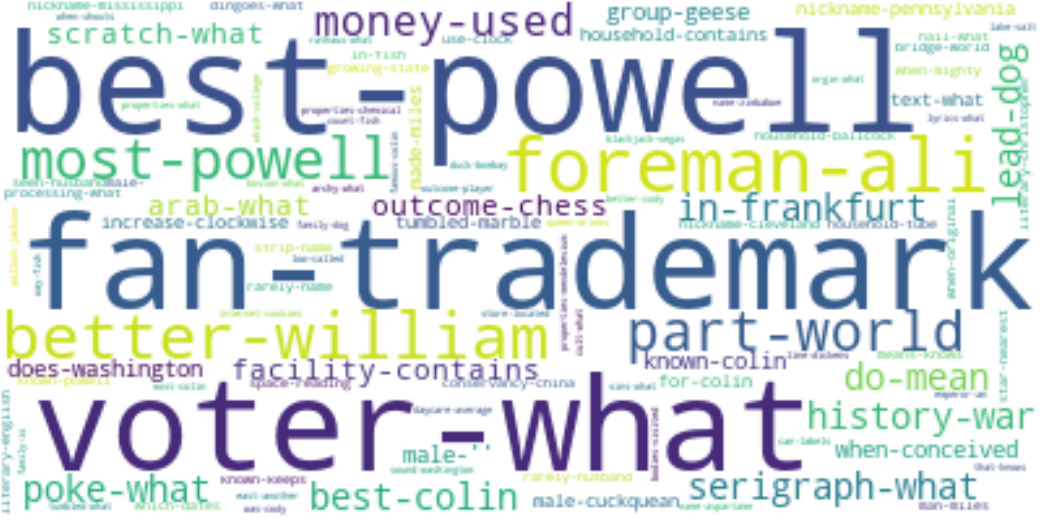}
    \end{minipage} \hspace{1mm}
    \begin{minipage}[t]{0.31\textwidth} 
      \centering 
      \includegraphics[width=1\linewidth]{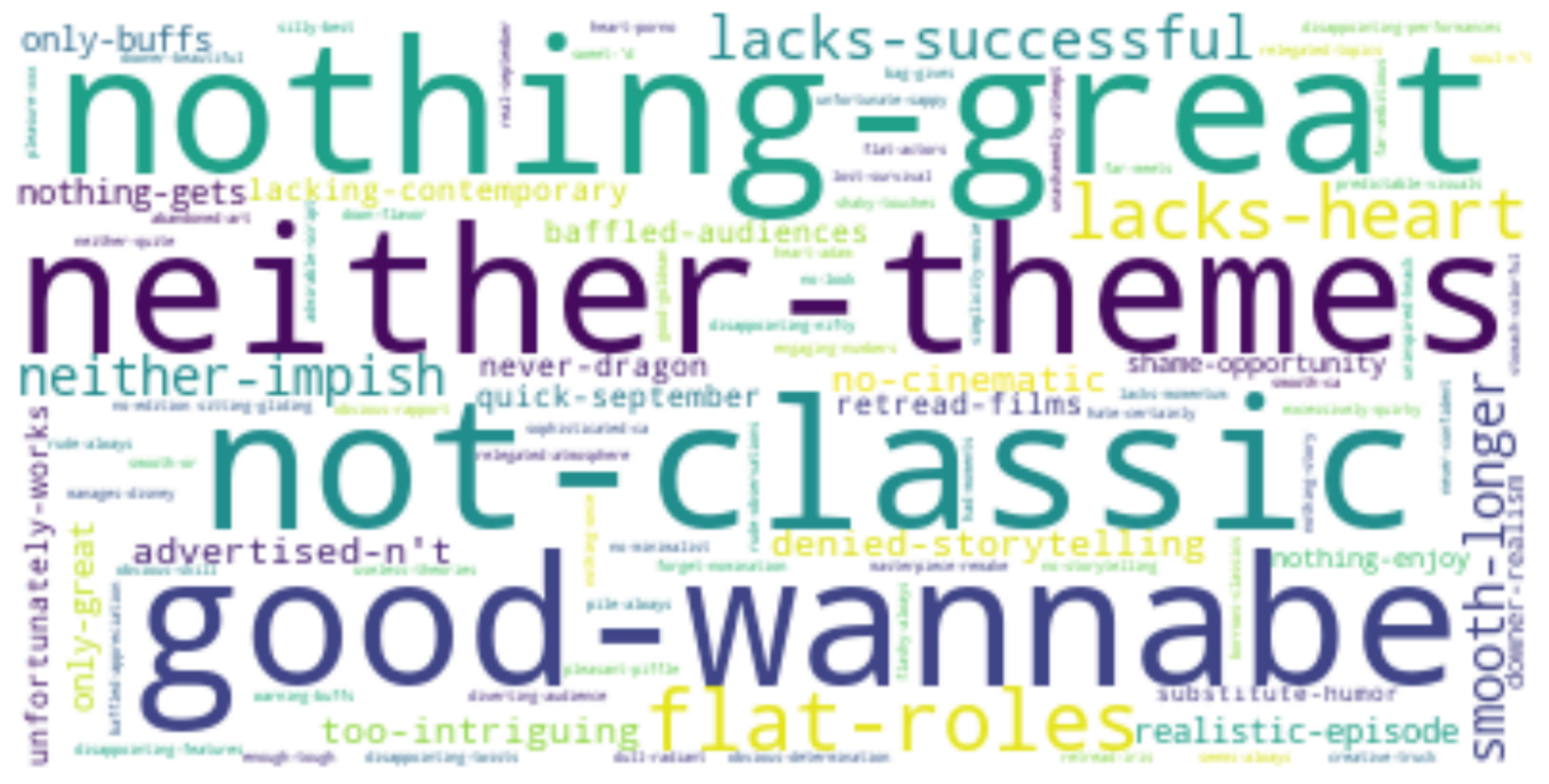}
\end{minipage} 
\begin{minipage}[t]{0.31\textwidth} 
      \centering 
      \includegraphics[width=1\linewidth]{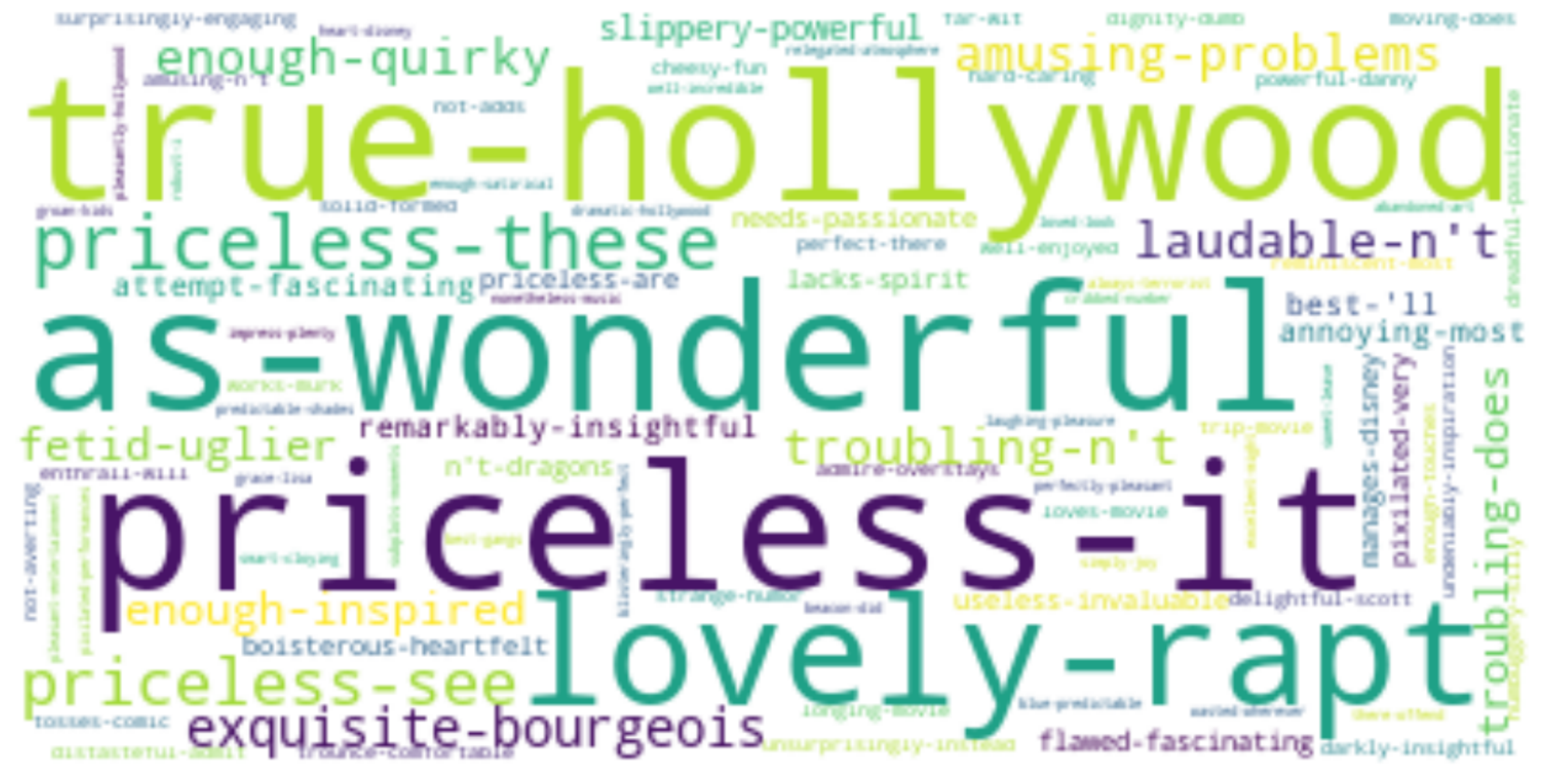}
    \end{minipage} 
       \caption{\label{fig:inter_cloud}Pairwise Word Interactions Cloud  on LSTM model for (left) TREC, (middle) SST-2, and (right) SST-1.}
\end{figure*}

\begin{figure*}[htbp] 
  \centering 
    \begin{minipage}[]{0.86\textwidth} 
      \centering 
      \includegraphics[width=1\columnwidth]{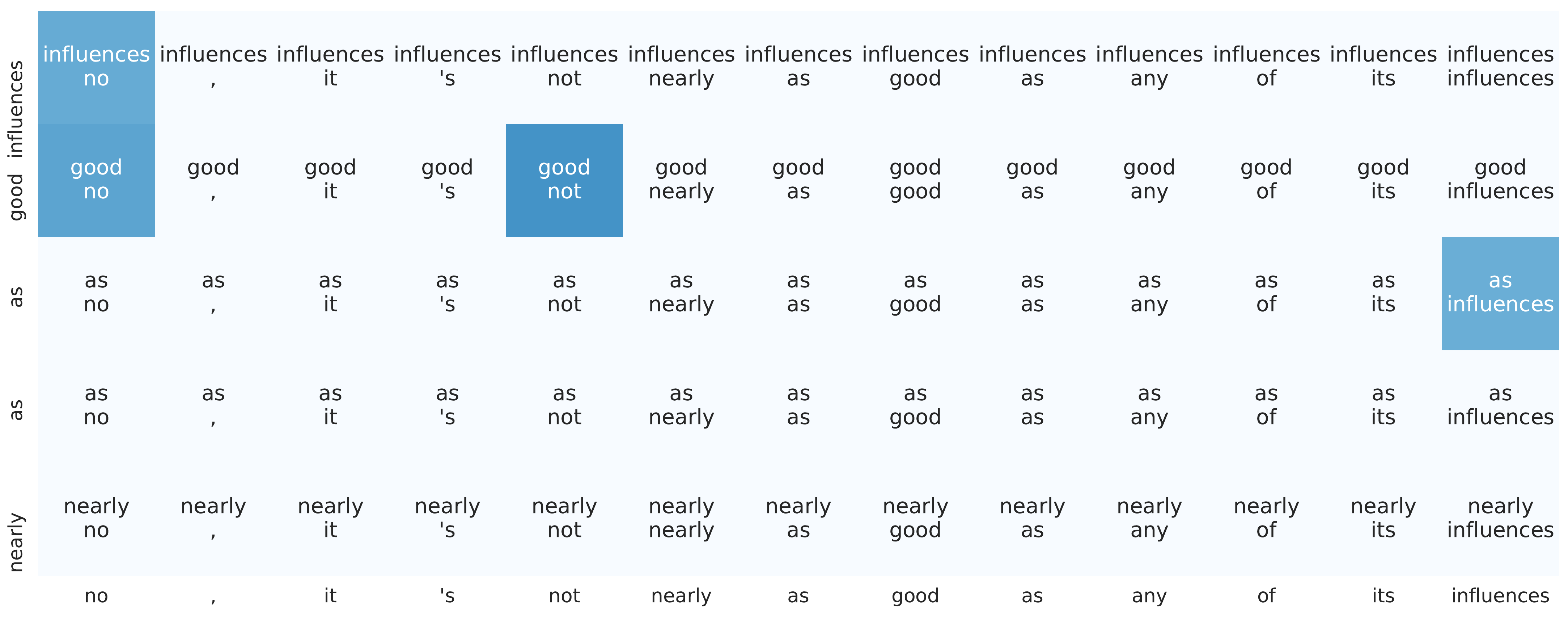}
      \label{fig:viz_not_4}
    \end{minipage} 
    \begin{minipage}[]{0.860\textwidth} 
      \centering 
      \includegraphics[width=1\columnwidth]{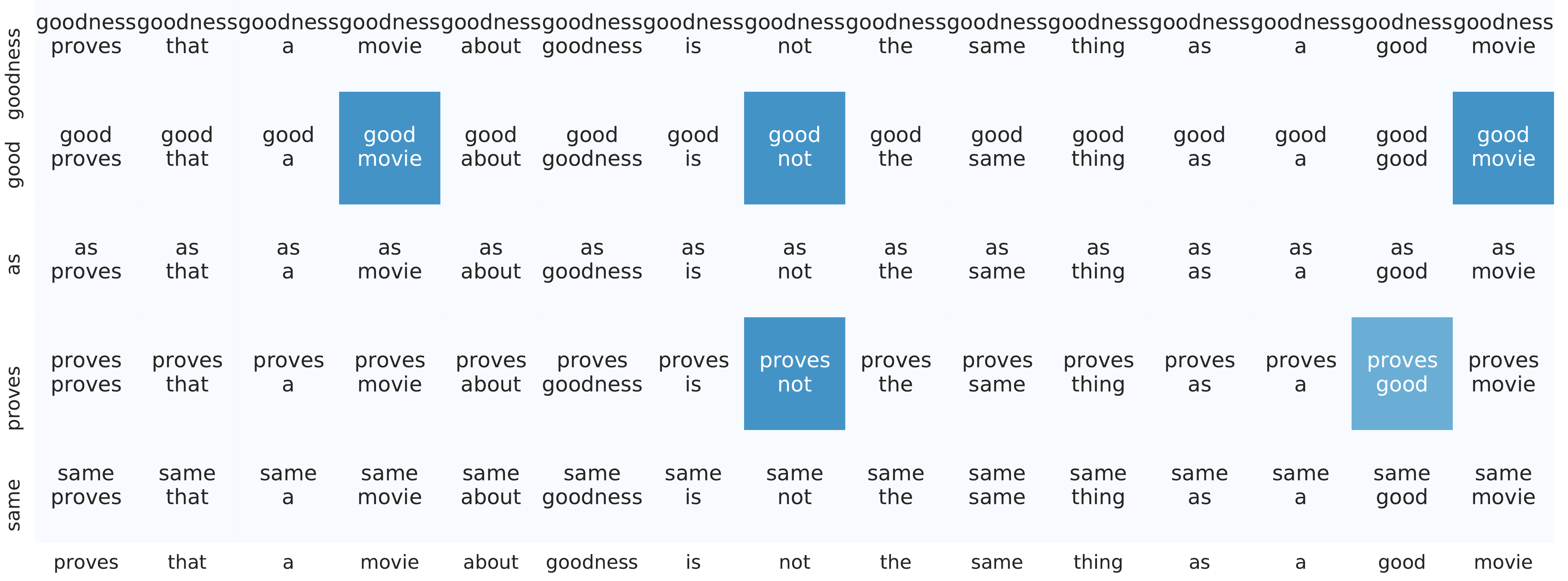}
      \label{fig:viz_not_5}
    \end{minipage} 
  \caption{ \label{fig:viz_not}Visualization of some learnt interactions viewed at the sentence level: For example, for the input sentence: "no, it's not nearly good as any of it's influences": "good" was picked out as an important word, "not" by itself was not an important word,  using $\mathbf{R}$, prediction using \methodName aggregates "good" with "not" in the sentence as 'not-good' is an important interaction globally, which is used to make a prediction. }
\end{figure*}

We show in Figure~\ref{fig:inter_freq_corr} that our learnt interaction matrix does not merely mirror the co-occurrence statistics, instead learns more general informative global interactions.   Figure~\ref{fig:inter_cloud} use word clouds to visualize interacted word pairs obtained from TREC, SST1 and SST2 datasets. The interactions are ranked based on $\mathbb{E}_q[\mathbf{A}_{x_{i,j}}| \x_{i,j}]$.  The selected word pairs are consistent with the corresponding tasks. 
 Figure~\ref{fig:viz_not} visualizes exemplar words' interactions that an improved model considers during inference.

In Table~\ref{tab:freq_corr}, we observe that the correlation between \methodName's word importance score $\mathbb{E}_q[\mathbf{R}_{x_t}| \x_{t}]$ and word frequency is lower than \vmask. This indicates \methodName can discover some important words that \vmask can not. %
 Table ~\ref{tab:freq_corr} also show a weak negative correlation between the interaction importance score$\mathbb{E}_q[\mathbf{A}_{x_{i,j}}| \x_{i,j}]$ and word co-occurrence frequency.

\begin{table}[th] 
    \begin{minipage}[]{0.5\textwidth} 
\centering
\scalebox{.81}{
   \begin{tabular}{|c|c|c|c|}\hline
        Dataset & \methodName-A-$R$   & \methodName-$\A$   & \vmask \\\hline
         SST2 & {\bf -0.0095} & -0.1182  & -0.0064\\\hline
         SST1 &  {\bf -0.0101}& -0.1059 & -0.0054 \\\hline
         SUBJ & {\bf -0.0137} & -0.0912 & -0.0077\\\hline
         TREC & {\bf -0.0139} & -0.1016 &-0.0118 \\\hline
    \end{tabular}}
    \end{minipage} 
\caption{{\bf Correlation with co occurrence statistics}: We summarize these observations using correlation statistics: \methodName has a lower correlation of the global word importance score $\mathbb{E}_q[x_t| \x_t]$ with word frequency(Column \methodName-$R$) than when it does not account for interactions (Column \vmask). We also show the correlation of the interaction importance scores $\mathbb{E}_q[A_{x_i,x_j}| \x_i,\x_j]$ with the co-occurrence frequency, which shows a weak negative correlation for all datasets(Column-\methodName-$\A$).  
    \label{tab:freq_corr}  } 
    \vspace{3mm}
\end{table}

\subsection{Detailed Derivation}
\label{sec:moreIBloss}

We follow the markov factorization : $p(\X,\Y,\bZ) = p(\bZ|\X,\Y) P(\X,\Y) = P(\bZ|\X,\Y) P(\Y|\X) P(\X) = P(\bZ|\X) P(\Y|\X) P(\X)$, motivated from the Markov assumption : $\Y \xleftrightarrow{} \X \xleftrightarrow{} \bZ$, i.e. $\Y$ and $\bZ$ are independent of each other given $\X$. We assume the data are generated using the above assumption where $\bZ$ is not observed, i.e. a latent variable.
Our derivation is based on \cite{chen2018learning,schulz2020restricting,alemi2016deep}, where we start from an approximation $q(\X,\Y,\bZ)$ instead of the true distribution $p(\X,\Y,\bZ)$. 
\paragraph{The lower bound for $I(\bZ;\Y)$.}
\begin{eqnarray}
I(\bZ,\Y)
&=& \sum_{\y, \z} q(\y,\z)\log \frac{q(\y,\z)}{q(\y)q(\z)}\nonumber\\
&=& \sum_{\y, \z} q(\y,\z)\log \frac{q(\y|\z)}{q(\y)}\nonumber\\
&=& \sum_{\y, \z} q(\y,\z) \log q(\y|\z)\nonumber\\
& & + H_q(\Y),
\end{eqnarray}
where $H_q(\cdot)$ represents entropy, and can be ignored for the purpose of training the model.

\begin{equation}
\begin{aligned}
\label{eq:L_bound}
& \sum_{\y, \z} q(\y,\z)\log q(\y|\z) \\
&= \sum_{\y, \z} q(\y,\z)\log \dfrac{q(\y|\z)p(\y|\z)}{p(\y|\z)} \\
&=\sum_{\y, \z} q(\y,\z)\log p(\y|\z) + KL[q(\y|\z)||p(\y|\z)] \\
&\geq \sum_{\y, \z} q(\y,\z)\log p(\y|\z),
\end{aligned}
\end{equation}
where $KL[\cdot||\cdot]$ denotes Kullback-Leibler divergence.
This gives us the following lower bound:
\begin{equation}
\begin{aligned}
\label{eq:izy-lower}
I(\bZ,\Y)
&\geq \sum_{\y, \z}q(\y,\z) \log p(\y|\z) + H_q(\y)\\
&=\sum_{\y, \z, \x}q(\x,\y,\z)\log p(\y|\z)+ H_q(\y)\\ 
&\!\!\!=\!\!\!\sum_{\y, \z, \x}\!\!q(\x, \y)q(\z|\x) \log p(\y|\z)\!+\!H_q(\y),
\end{aligned}
\end{equation}
where the last step uses $q(\x,\y,\z) = q(\x)q(\y|\x) q(\z|\x)$, which is a factorization based on the conditional dependency: $\y \leftrightarrow \x \leftrightarrow \z$: $\y$ and $\z$ are independent given $\x$. 

Given a sample, $(\x^{(m)},\y^{(m)})$, we can assume the empirical distribution $q(\x^{(m)},\y^{(m)})$  simply defined as a multiplication of two Delta functions
\begin{equation}
q(\x=\x^{(m)},\y=\y^{(m)}) = \delta_{\x^{(m)}}(\x)\cdot\delta_{\y^{(m)}}(\y).
\end{equation}
So we simplifying further of the first term:
\begin{equation}
\begin{aligned}
\label{eq:vib-1}
I(\z;\y^{(m)})
&\geq \sum_{\z}q(\z|\x^{(m)}) \log p(\y^{(m)}|\z)\\
&=\mathbb{E}_{q(\z|\x^{(m)})} \log(p(\mathbf{y}^{(m)}| \z))
\end{aligned}
\end{equation}
Since $\mathbf{Z} = \diag{\R_{\x}} \mathbf{E'}$, and $\mathbf{E'}$ is a deterministic function of $\mathbf{A}$, we have:
\begin{equation}
\begin{aligned}
\label{eq:vib-1_1}
I(\z;\y^{(m)}) \geq \mathbb{E}_{q(\mathbf{z}|\x^{(m)})} log (p(\mathbf{y}^{(m)}| \mathbf{R}, \mathbf{A}, \x^{(m)}))
\end{aligned}
\end{equation}

\paragraph{The upper bound for $I(\bZ;\X)$.}
\begin{equation}
\begin{aligned}
I(\bZ,\X)
&=& \sum_{\x, \z} q(\x,\z)\log \frac{q(\x,\z)}{q(\x) q(\z)}\nonumber\\
&=& \sum_{\x, \z} q(\x,\z)\log \frac{q(\z|\x)}{q(\z)}\nonumber
\end{aligned}
\end{equation}

\begin{equation}
\begin{aligned}
&=& \sum_{\x, \z} q(\x,\z)\log q(\z|\x)\nonumber\\
& & - \sum_{\x, \z} q(\x,\z)\log q(\z)
\end{aligned}
\end{equation}

By replacing $q(\z)$ with a prior distribution of $\z$, $p_0(\z)$, we have
\begin{equation}
\sum_{\x, \z} q(\x,\z)\log q(\z) \geq \sum_{\x, \z} q(\x,\z)\log p_0(\z).
\end{equation}
Then we can obtain an upper bound of the mutual information
\begin{eqnarray}
\label{eq:vib-2}
I(\bZ;\X)
&\leq& \sum_{\x, \z} q(\x,\z)\log q(\z|\x)\nonumber\\
& & - \sum_{\x, \z}q(\x,\z)\log p_0(\z)\nonumber\\
&=& \sum_{\x} q(\x)KL[q(\z|\x)||p_0(\z)]\nonumber\\
&=&  \mathbb{E}_{q(x)}{KL[q(\z|\x)||p_0(\z)}].
\end{eqnarray}
For a given sample $(\x^m,\y^m)$, 
\begin{equation}
I(\z;\x^{(m)}) = KL[q(\z|\x^{(m)})||p_0(\z)]
\end{equation}

Given $\mathbf{X}$,  we assume $\mathbf{R}$ and $\mathbf{A}$ are independent. We also assume a factorable prior $p_0( \z) = p_{r0}(\mathbf{R})p_{a0}(\mathbf{A})$. This gives us:
\begin{eqnarray}
\label{eq:vib-2_1}
I(\z,\x^{(m)})\leq  KL(q(\mathbf{R} | \x^{(m)}) || p_{r0}(\mathbf{R}))\nonumber \\
   + KL(q(\mathbf{A} | \x^{(m)}) || p_{a0}(\mathbf{A}))
\end{eqnarray}

\label{sec:imask_vis}
\begin{figure*}
    \centering
    \includegraphics[width=\textwidth]{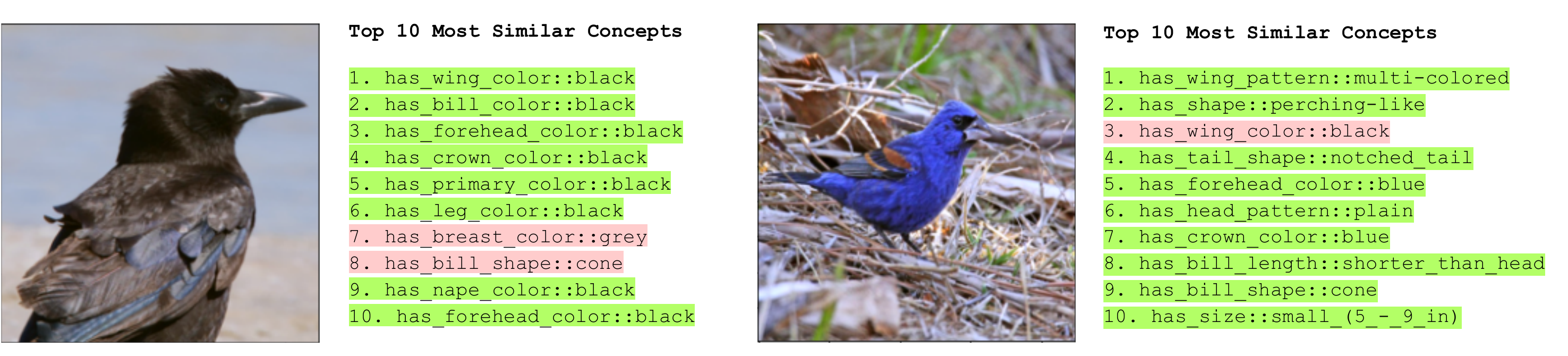}
    \caption{Qualitative assessment of the \method layer. We find the most similar concepts to a given image by sorting them using cosine distances between the image and concept embeddings. Ground truth concepts are highlighted in green.}
    \label{fig:imask_vis}
\end{figure*}

\section{Extending  \method for Capturing Concept Interactions in Vision Tasks}
\label{sec:conceptMore}
In this section, we demonstrate the application of the \method layer in the context of image modeling with convolutional neural networks. 
\subsection{Dataset and Concept Pre-processing}
We train a joint concept bottleneck~\cite{koh2020concept} model on the CUB dataset~\cite{wah2011caltech} which comprises of 11,788 photographs of birds from 200 species, where each image has additional annotations of 312 binary concepts corresponding to bird attributes like wing color, beak shape, etc. The concept annotations in the CUB dataset are noisy, hence we adopt the method described in concept bottleneck models~\cite{koh2020concept} to pre-process the concept level annotations and remove noisy labels. Each concept annotation was provided by a single crowdworker who is not an avian expert, and the concepts can be quite similar to each other, e.g., some annotators might say a bird has a red belly, while other annotators might say that the belly is rufous (reddish-brown) instead. In order to deal with such issues, we aggregate instance-level concept annotations into class-level concepts via majority voting, i.e if more than 50\% of crows have black wings in the data, then the class crow is considered to always have black wings. As a result, images with the same class have the same concept annotations. While this approximation is mostly true for this dataset, there are some exceptions due to visual occlusion, as well as sexual and age dimorphism. After majority voting, we further filter out concepts that are too sparse, keeping only concepts (binary attributes) that are present after majority voting in at least 10 classes. After this filtering, we are left with 112 concepts.

\subsection{Method}
\label{sec:conceptMethod}

Given an image $x \in \mathbb{R}^d$, its class label $y$, and concept labels $c$, which is a vector of $k$ concepts; we train a concept bottleneck model with a \method layer in order to model the interactions between the $k$ concepts. Additionally, we encode concepts as learned high-dimensional representations in a shared image-concept latent space, such that a dynamic interaction graph $\mathbf{A}'$ can be defined with concepts and the image as neighboring nodes. Next we use the graph convolutional operation step to update the representation of each node with its neighbors. The resulting node representations are max-pooled and the final image-concept representation is used for task classification. Overall, we learn a global interaction matrix $\mathbf{A} = \{\mathbf{A}\}_{k \times k}$ between concepts, we use the learnt graph and concept representations to train a task prediction head for the final task.

Consider an image encoder model, $f: \mathbb{R}^d \to \mathbb{R}^k$, that encodes the image into a d-dimensional representation, $f(x)$. This representation is then transformed through a shallow MLP, $g: \mathbb{R}^d \to \mathbb{R}^k$ which maps the image from the representation space to the concept space to give $g(f(x))$. This is then passed through a concept prediction head layer and treated as a mutli-label classification problem for predicting concepts. We also define a  MLP $h: \mathbb{R}^k \to \mathbb{R}$ which finally maps the representation from the concept to the task prediction space which is modeled as a multi-class classification problem. Now, in order to adapt the \method layer for learning concept interactions, we need to formulate these concepts as high-dimensional representations in the image embedding space, such that the \method layer can used to treat the concept embeddings as nodes and their interactions as the edges in an interaction graph. As such we define a randomly initialized embedding layer that helps us query concept representations given concept labels $c$. We make the assumption that the latent space for the concept representation is aligned with that of image representations. Intuitively, we are assuming that the image itself can be represented as a composition of different concept representations and hence can be embedded in the same space. This allows us to use the image representation $f(x)$ as another node in the $\mathbf{A}$ graph during the subsequent graph convolution update step. Therefore during training, we dynamically compute the interaction between the image and the concept representations as their cosine similarity scaled between $0$ and $1$. These interactions are then used to expand the graph to give $\mathbf{A'} = \{\mathbf{A'}\}_{(k+1) \times (k+1)}$ where each $\mathbf{A'}_{ij} \in \{0,1\}$ with image as the added node and the image-concept scaled similarities as the corresponding interactions in the original graph $A$. The dynamic interaction graph $\mathbf{A'}$, image representation $f(x)$, and concept embeddings $X_c$ are transformed using the graph convolution update described in section \ref{sec:gcn} as,
$\mathbf{E'}=g_{GC}(\mathbf{X'},\mathbf{A'})$, where $\mathbf{X'} = [f(x), X_c]$ is a concatenation of the image and concept representations corresponding to the interaction graph $A'$. Output representations $E'$ are max-pooled and passed to the prediction head $p: \mathbb{R}^d \to \mathbb{R}$. The classification loss at the prediction head is used to train the \method layer.

Overall, for each data point $\{x, y, c\}$, we get the image representation $f(x)$ from the CNN encoder, and corresponding concept representations for each $c_i$ from the embedding layer. The interactions are computed between $f(x)$ and each concept representation and the image itself is added to the interaction graph as the neighbor of the observed concepts, $c$. Similar to previous formulations, we intend to learn the unknown graph $\mathbf{A} = \{\mathbf{A}\}_{V \times V}$ where each $\mathbf{A}_{ij} \in \{0,1\}$ specifies the strength of the interaction. Through this task we also learn concept embeddings in the same latent space as the image embeddings from the CNN encoder.

\subsection{Results}
We use a concept bottleneck models~\cite{koh2020concept} as a baseline to demonstrate the applicability of \method for image modeling scenarios. Specifically, we train an Inception-V3 model with a concept bottleneck for the bird identification task on the Caltech-UCSD Birds-200-2011 (CUB) dataset~\cite{wah2011caltech}. With the model, we jointly train a \method layer to learn the interactions between the concepts. \method also learns concept embeddings in a joint image-concept space. In order to demonstrate this, we use the image representation to find most similar concepts using the cosine metric. In Figure~\ref{fig:imask_vis}, we show that the model is able to encode the image very close to the ground-truth concept embeddings. Additionally, as demonstrated in~\citet{koh2020concept}, the model is able to predict concepts very accurately at the bottleneck while maintaining a final bird classification task accuracy of $80.04\%$. Note that this allows us to reasonably intervene on the concepts. With test-time intervention, the model is able to improve its performance by $+9.37\%$ to $89.41\%$. Similar to this, \method when used with the CNN's image representation and intervened concepts is able to achieve a prediction accuracy of $95.74\%(+15.70\%)$ at its prediction head.

\end{document}